\documentclass{ecai} 

\usepackage{latexsym}
\usepackage{amssymb}
\usepackage{amsmath}
\usepackage{amsthm}
\usepackage{booktabs}
\usepackage{enumitem}

\usepackage{soul}
\usepackage{url}
\usepackage[hidelinks]{hyperref}
\usepackage[utf8]{inputenc}
\usepackage[small]{caption}
\usepackage{graphicx}
\usepackage{booktabs}

\usepackage{algorithm}
\usepackage{algorithmic}
\usepackage{array}
\usepackage{multirow}
\usepackage{xspace}
\usepackage[dvipsnames]{xcolor}
\usepackage{pifont}

\definecolor{myRed}{rgb}{0.808,0.067,0.149}
\definecolor{myGreen}{rgb}{0.067,0.708,0.149}

\DeclareRobustCommand\onedot{\futurelet\@let@token\@onedot}
\def\@onedot{\ifx\@let@token.\else.\null\fi\xspace}

\newcommand{\ie}{\emph{i.e.}}
\newcommand{\eg}{\emph{e.g.}}
\newcommand{\etal}{\emph{et al.}}
\newcolumntype{P}[1]{>{\centering\arraybackslash}p{#1}}
\newcolumntype{M}[1]{>{\centering\arraybackslash}m{#1}}

\newcommand{\xmark}{{\color{myRed}\ding{55}}}%
\newcommand{\cmark}{{\color{myGreen}\ding{51}}}

\newcommand{\BibTeX}{B\kern-.05em{\sc i\kern-.025em b}\kern-.08em\TeX}

\begin{document}


\begin{frontmatter}


\paperid{764} 


\title{CBM: Curriculum by Masking}


\author[A]{\fnms{Andrei}~\snm{Jarc\u{a}}\footnote{Equal contribution.}}
\author[A]{\fnms{Florinel-Alin}~\snm{Croitoru}\footnotemark}
\author[A]{\fnms{Radu Tudor}~\snm{Ionescu}\thanks{Corresponding Author. Email: raducu.ionescu@gmail.com.}} 

\address[A]{Department of Computer Science, University of Bucharest, 14 Academiei, Bucharest, Romania}


\begin{abstract}
We propose Curriculum by Masking (CBM), a novel state-of-the-art curriculum learning strategy that effectively creates an easy-to-hard training schedule via patch (token) masking, offering significant accuracy improvements over the conventional training regime and previous curriculum learning (CL) methods. CBM leverages gradient magnitudes to prioritize the masking of salient image regions via a novel masking algorithm and a novel masking block. Our approach enables controlling sample difficulty via the patch masking ratio, generating an effective easy-to-hard curriculum by gradually introducing harder samples as training progresses. CBM operates with two easily configurable parameters, \ie~the number of patches and the curriculum schedule, making it a versatile curriculum learning approach for object recognition and detection. We conduct experiments with various neural architectures, ranging from convolutional networks to vision transformers, on five benchmark data sets (CIFAR-10, CIFAR-100, ImageNet, Food-101 and PASCAL VOC), to compare CBM with conventional as well as curriculum-based training regimes. Our results reveal the superiority of our strategy compared with the state-of-the-art curriculum learning regimes. We also observe improvements in transfer learning contexts, where CBM surpasses previous work by considerable margins in terms of accuracy. We release our code for free non-commercial use at \url{https://github.com/CroitoruAlin/CBM}.
\end{abstract}

\end{frontmatter}


\setlength{\abovedisplayskip}{3.0pt}
\setlength{\belowdisplayskip}{3.0pt}

\section{Introduction}

\begin{figure*}[!t]
\begin{center}
\centerline{\includegraphics[width=0.72\linewidth]{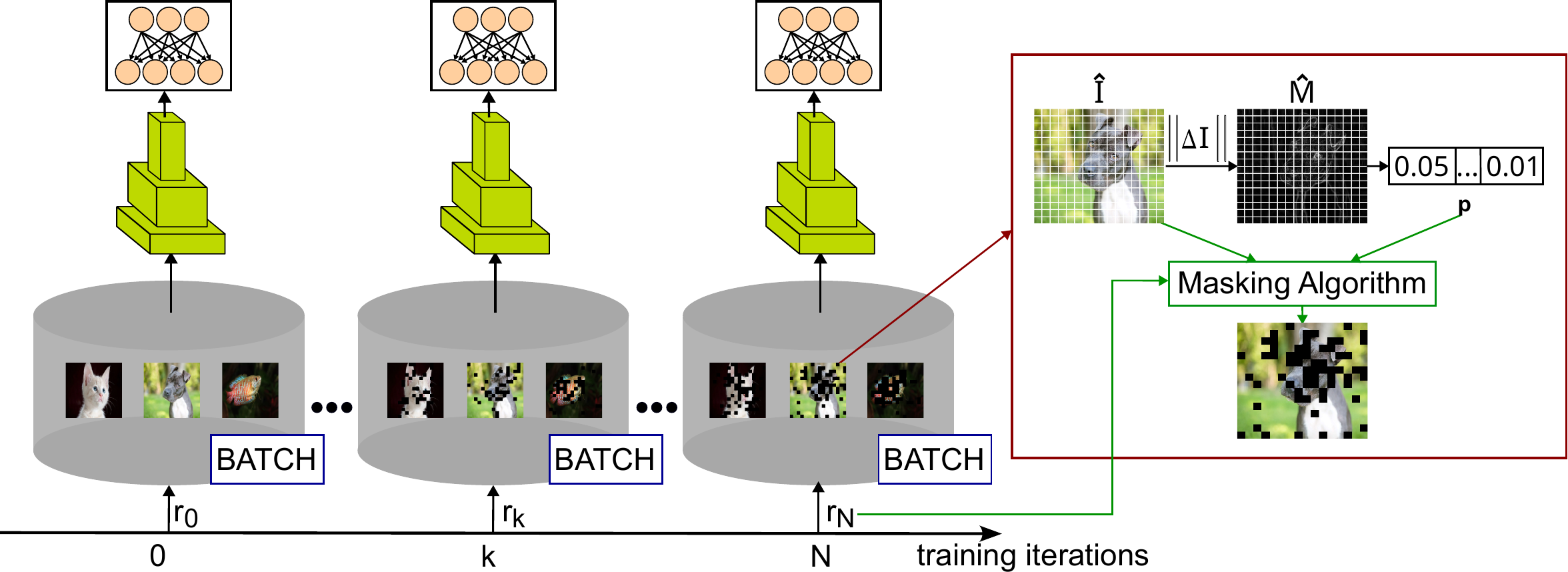}}
\vspace{0.1cm}
\caption{An overview of Curriculum by Masking. The training starts with fully visible images. During training, the patch masking ratio is gradually increased to make the samples more difficult. The masking is predominantly focused on the more salient regions (with higher gradient magnitudes), to reduce the likelihood of producing easier images by masking the background information. Best viewed in color.}
\label{fig_cbm}
\vspace{-0.3cm}
\end{center}
\end{figure*}

Humans learn by grasping the easier concepts before gradually moving to the more complex ones. Inspired by this observation, Bengio \etal~\citep{Bengio-ICML-2009} proposed curriculum learning, a training regime that introduces a structured order of the data samples to train neural models from easy to hard. This method ensures that the examples are presented to the model in a logical and meaningful sequence, allowing the model to effectively learn and develop its knowledge base. The method has been employed in multiple scenarios yielding significant performance improvements \citep{Soviany-IJCV-2022}. Furthermore, over the course of time, various methods have been developed to integrate curriculum learning \citep{Bengio-ICML-2009,Chen-ICCV-2015,Croitoru-arXiv-2022,Gui-FG-2017,Jiang-ICML-2018,Kocmi-RANLP-2017,Liu-IJCAI-2018,Madan-WACV-2024,Platanios-NAACL-2019,Shi-ECCV-2016,Sinha-NIPS-2020,Soviany-CVIU-2021}. While many of these approaches involve different components of the training process, Soviany \etal~\citep{Soviany-IJCV-2022} have classified them into four main categories. The first category, known as data-level curriculum, involves sorting the data samples based on certain difficulty criteria. This approach aligns with the initial implementation of Bengio \etal~\citep{Bengio-ICML-2009}. The second category, referred to as model-level curriculum, includes methods that gradually increase the capacity of the model as the training progresses. The third category, represented by task-level curriculum, aims to make the learning task more intricate over time. Lastly, objective-level curriculum begins with a simplified objective, \eg~a convex objective, and transforms it during training until it becomes the final target objective, which is usually non-convex. 

The data-level curriculum learning approaches involve sorting the training examples based on some difficulty metric \citep{Bengio-ICML-2009,Ionescu-CVPR-2016,Kocmi-RANLP-2017,Pentina-CVPR-2015,Soviany-CVIU-2021,Tay-ACL-2019,Wei-WACV-2021,Zhang-ISPASS-2021}, before proceeding with the actual learning process. However, this approach has a significant challenge associated with it, namely the need to use a custom difficulty metric for each domain. The choice of the difficulty metric may vary depending on the specific learning task, and, in certain cases, it can be very hard to propose a useful difficulty metric \citep{Croitoru-arXiv-2022}. Building on the success of self-supervised approaches \citep{He-CVPR-2022,Devlin-NAACL-2019} used to train deep learning models to reconstruct masked information (tokens), we propose a novel data-level curriculum learning approach, termed Curriculum by Masking (CBM), that artificially raises the difficulty level of each training image by masking a certain number of patches, where the number of masked patches gradually increases during the training process, as illustrated in Figure~\ref{fig_cbm}. Hence, our approach does not require the prior sorting of data samples, as it enables full control over the complexity of a data sample via the masking ratio. However, we conjecture that randomly masking patches does not necessarily induce an easy-to-hard curriculum. For example, if the masked patches happen to hide most of the background in an image, leaving the foreground object that needs to be recognized or detected visible, we will actually end up with an easier example instead of a more difficult one. To this end, we propose to sample the masked patches from a probability distribution derived from image gradient magnitudes, essentially prioritizing the masking of salient image patches. Since salient regions are more likely to contain discriminative patterns, masking patches with larger gradient magnitudes reduces the number of visible discriminative patterns, thus increasing the difficulty of recognizing objects in images.

We conduct experiments to determine the effectiveness of our approach in object recognition and object detection. We apply CBM on different types of convolutional and transformer neural networks, \eg~ResNet-18 \citep{He-CVPR-2016}, Wide-ResNet-50 \citep{Zagoruyko-ArXiv-2016wide}, CvT-13 \citep{Wu-ICCV-2021}, and YOLOv5 \citep{Jocher-zenodo-2022}. Our empirical study is carried out on five data sets: CIFAR-10 \citep{Krizhevsky-TECHREP-2009}, CIFAR-100 \citep{Krizhevsky-TECHREP-2009}, ImageNet \citep{Russakovsky-IJCV-2015}, Food-101 \citep{Bossard-ECCV-2014} and PASCAL VOC \citep{Everingham-IJCV-2010}. We compare our approach with four state-of-the-art curriculum learning methods \citep{Croitoru-arXiv-2022,Dogan-ECCV-2020,Sinha-NIPS-2020,Wang-ICCV-2023}, as well as two more baseline training regimes. The first baseline is the vanilla training regime, while the second one employs the vanilla patch masking technique proposed by He \etal~\citep{He-CVPR-2022}. Lastly, we present ablation results and a comprehensive comparison of masking ratio schedules, identifying multiple successful configurations and parameter choices for our method.

In summary, our contribution is threefold:
\begin{itemize}
    \item We propose a novel curriculum learning method based on masking an increasingly higher number of patches during training.
    \item We introduce a patch selection strategy that prioritizes the masking of patches with larger gradient magnitudes, ensuring an easy-to-hard curriculum. 
    \item We empirically demonstrate the effectiveness of CBM in object recognition and object detection for multiple neural architectures, comparing it with several competing training regimes.
\end{itemize}

\section{Related Work}

Curriculum learning is a training technique introduced by Bengio \etal~\citep{Bengio-ICML-2009}, which provides the training examples in a meaningful order, from easy to hard, to neural networks. The objective is to enhance the performance of neural models, while also improving the convergence speed of the training process. Since its introduction, curriculum learning has proven its effectiveness in various domains, such as computer vision \citep{Bengio-ICML-2009,Chen-ICCV-2015,Croitoru-arXiv-2022,Gui-FG-2017,Jiang-ICML-2018,Shi-ECCV-2016,Sinha-NIPS-2020,Soviany-CVIU-2021}, natural language processing \citep{Bengio-ICML-2009,Croitoru-arXiv-2022,Kocmi-RANLP-2017,Liu-IJCAI-2018,Platanios-NAACL-2019,Spitkovsky-NIPS-2009}, and signal processing \citep{Amodei-ICML-2016,Croitoru-arXiv-2022,Ranjan-ACM-2018}. The method has been very successful and has undergone extensive development, as illustrated in some recent surveys \citep{Soviany-IJCV-2022,Wang-PAMI-2021}. These developments range from strategies for measuring data difficulty \citep{Bengio-ICML-2009,Ionescu-CVPR-2016,Kocmi-RANLP-2017,Pentina-CVPR-2015,Shi-ECCV-2016,Soviany-CVIU-2021,Tay-ACL-2019,Wei-WACV-2021,Zhang-ISPASS-2021} to methods focusing on other aspects of the training process \citep{Burduja-ICIP-2021,Croitoru-arXiv-2022,Caubriere-INTERSPEECH-2019,Karras-ICLR-2018,Sinha-NIPS-2020}.
A well-known method to apply curriculum learning is by defining a metric that evaluates the complexity of the data, and subsequently arranging the training examples from the simplest to the most challenging ones based on the respective metric. Researchers have made significant strides in finding improved metrics for various domains and tasks. For instance, images containing fewer and larger objects in computer vision are deemed easier than other images \citep{Shi-ECCV-2016,Soviany-CVIU-2021}. In natural language processing, word frequency  \citep{Bengio-ICML-2009,Liu-IJCAI-2018} and sequence length \citep{Cirik-Arxiv-2016,Kocmi-RANLP-2017,Tay-ACL-2019,Zhang-ISPASS-2021} are utilized to assess the sample difficulty. In some cases, researchers have also integrated human feedback into their metric design \citep{Sanchez-MICCAI-2019,Pentina-CVPR-2015,Wei-WACV-2021}.

The aforementioned curriculum strategies have proven to be effective. However, they have been found to lack practicality due to their reliance on human expert input \citep{Sanchez-MICCAI-2019}, which may not always be available. Moreover, these methods remain fixed during training and may not adapt the curriculum to the changing needs of the models. As a result, the research community developed new curriculum-based approaches to overcome these limitations. For instance, Kumar \etal~\citep{Kumar-ANIPS-2010} introduced self-paced learning, a method that measures the difficulty of the training samples based on the performance of the trained model. Thus, the order of the training samples changes according to the model feedback during training, and thanks to this property, several works adopted the approach \citep{Fan-AAAI-2017,Gong-TEVC-2019,Jiang-AAAI-2015,Kumar-ANIPS-2010,Li-AAAI-2016,Ristea-INTERSPEECH-2021,Jhou-PR-2018}. Moreover, it is possible to implement a combination of self-paced learning and classic curriculum learning approaches. This approach has been previously utilized under the name of self-paced curriculum learning \citep{Jiang-AAAI-2015,Soviany-IJCV-2022}. Another popular method is teacher-student curriculum learning \citep{Hacohen-ICML-2019,Wu-NIPS-2018,Zhang-AS-2019}, where the teacher learns to supervise the student network via a curriculum.

Methods that fall under the model-level curriculum learning paradigm \citep{Burduja-ICIP-2021,Croitoru-arXiv-2022,Karras-ICLR-2018,Sinha-NIPS-2020,Soviany-IJCV-2022} are closer to our work. In this setting, the curriculum does not imply ordering the samples in ascending order of their difficulty. Instead, the curriculum is implemented by increasing the model capacity as the training progresses, or by adjusting the task to become more accessible at the beginning of the training.  Curriculum by Smoothing (CBS), a technique developed by Sinha \etal~\citep{Sinha-NIPS-2020}, blurs the activation maps resulting from convolutional layers during the training process to let the model focus on the bigger picture. CBS gradually reduces the amount of blur as the model improves. This approach has been successful on various data sets and models. However, it does require extra processing steps during training, which can make the learning process last longer. The work of Burduja \etal~\citep{Burduja-ICIP-2021} is an alternative to CBS, where the input images are blurred instead of the intermediary activation maps.
Another example is Learning Rate Curriculum (LeRaC) \citep{Croitoru-arXiv-2022}. This method assigns different learning rates to the network layers based on their proximity to the input. Layers closer to the input have higher learning rates, while those farther away have lower rates. Over time, the learning rates are adjusted to converge to a consistent value. 
Similar to \citep{Burduja-ICIP-2021,Croitoru-arXiv-2022,Sinha-NIPS-2020}, our approach does not require an external difficulty measure. 

Different from other curriculum learning approaches, we propose to induce a curriculum via a progressive masking of input patches. To the best of our knowledge, we are the first to introduce a curriculum learning method based on patch masking. Furthermore, we go beyond a naive implementation and apply the masking operation by taking into consideration the salience (gradient magnitude) of the patches to create the premises for an easy-to-hard curriculum. This differs significantly from other approaches \citep{Burduja-ICIP-2021,Sinha-NIPS-2020} which apply the smoothing operation uniformly in space, using a single filter applied at every location of the input via convolution. Another advantage in favor of CBS \citep{Sinha-NIPS-2020} is that our approach does not include auxiliary operations after each neural layer of the model, so the training time is identical to that of the conventional training regime. 


\section{Method}

Masking specific parts, \eg~patches or tokens, of input data samples has been demonstrated to be a successful technique to train deep learning models in a self-supervised manner, in both natural language processing \citep{Devlin-NAACL-2019,Li-CVPR-2023} and computer vision \citep{Bandara-CVPR-2023,Feichtenhofer-NeurIPS-2022,Georgescu-arXiv-2022,He-CVPR-2022,Li-CVPR-2023,Tong-NeurIPS-2022} domains. This method has been primarily used as a pre-training task \citep{Devlin-NAACL-2019,He-CVPR-2022}, in which the model is tasked with reconstructing the masked information. Through this approach, the model is able to learn and recognize patterns in the data, leading to improved accuracy and more effective performance in downstream tasks. In this work, we redesign the masking procedure to create a curriculum learning method based on masking patches according to their salience level. Our approach is able to artificially generate examples of various difficulty levels. We name the resulting learning procedure Curriculum by Masking (CBM). 

Our curriculum learning procedure starts from the original samples and gradually creates more difficult images as the training progresses. We increase the image difficulty by masking a higher number of heterogeneous patches. We estimate the heterogeneity (or salience) of a patch via the average gradient magnitude computed on both horizontal and vertical axes. Such patches are likely to contain regions of interest, such as object parts or other discriminative patterns. Before the training starts, our approach creates a curriculum schedule vector $\mathbf{r} \in \mathbb{R}^N$, where each element $r_k$ represents the percentage of patches that are to be masked during the $k$-th training iteration. We note that the maximum number of iterations is denoted by $N$, and $r_N$ is equal to a maximum masking ratio, which is fixed beforehand through validation. We empirically study various alternative functions to generate the curriculum schedule $\mathbf{r}$ and regulate its growth rate. For details about the results obtained with different curriculum schedules, please refer to Section~\ref{section_exp}.

\begin{algorithm}[t]
\caption{Salience-Based Masking Algorithm\label{alg_masking}}
\label{alg:algorithm}
\textbf{Input}: 
$r_k$ -- the percentage of masked patches at iteration $k$;

$\hat{I}$ -- the input image;

$n$ -- the number of patches in an image;

$\mathbf{p}$ -- an array of size $n$ containing the masking probabilities for each patch.\\
\textbf{Output}: $\hat{I}_{mask}$ - the masked image.\\
\textbf{Computation}:
\begin{algorithmic}[1] 
\STATE $n_{mask} \leftarrow n \cdot r_k$
\STATE $\hat{I}_{mask} \leftarrow \hat{I}$
\FOR{$i$ in $1\dots n_{mask}$}
\STATE $j\sim$ Categorical$(n, \mathbf{p})$
\STATE $\hat{I}_{mask}[j] \leftarrow 0$
\ENDFOR
\STATE \textbf{return} $\hat{I}_{mask}$
\end{algorithmic}
\end{algorithm}

Our salience-based masking procedure is described in Algorithm~\ref{alg_masking}. The algorithm operates on the input image $\hat{I}$, which is divided into non-overlapping patches and represented as a tensor of dimensions $\mathbb{R}^{n \times \hat{h} \times \hat{w} \times c}$, where $n$ denotes the number of patches, $\hat{h}$ and $\hat{w}$ refer to the height and width of each patch, and $c$ represents the number of color channels. In addition to the input image, the algorithm also relies on the ratio of masked patches $r_k$ for the current training iteration $k$, and an array of probabilities $\mathbf{p} \in [0,1]^n$ that controls the likelihood of masking each patch. 

First, the algorithm computes the number of patches to be masked, denoted as $n_{mask}$, based on the specified percentage $r_k$. Then, at each loop iteration $i$, the algorithm samples a patch index $j$ from the set $\{1, \dots, n\}$ according to a categorical distribution described by the vector $\mathbf{p}$. Then, the patch at index $j$ is masked by setting its pixels to zero in the output image $\hat{I}_{mask}$.

As previously stated, the masking algorithm prioritizes salient regions of the input image. This property is accomplished by controlling the probabilities in $\mathbf{p}$, such that salient patches (likely to contain discriminative patterns) are assigned with higher probabilities, thus increasing their chances of being selected for masking. We hypothesize that salient regions can be identified to a certain extent by analyzing the magnitudes of the image gradients. We transform color images to the grayscale, before computing the gradients. Formally, given a grayscale image $I \in \mathbb{R}^{h \times w}$, where $h = n_h \cdot \hat{h}$, $w = n_w \cdot \hat{w}$ and $n = n_h \cdot n_w$, we compute its gradient as follows:
\begin{equation}
    \label{eq_grad}
    \Delta I = \left[\frac{\partial I}{\partial x}, \frac{\partial I}{\partial y} \right].
\end{equation}
Then, we computed the magnitude $M \in \mathbb{R}^{h \times w}$ of the gradient as follows:
\begin{equation}
    \label{eq_magnitude}
    M = \|\Delta I\| = \sqrt{\frac{\partial I}{\partial x}^2 +
   \frac{\partial I}{\partial y}^2 }.
\end{equation}
Next, we split $M$ into $n$ patches and obtain the tensor denoted by $\hat{M} \in \mathbb{R}^{n \times \hat{h} \times \hat{w}}$. Finally, we compute the elements of $\mathbf{p}$ via the following equations:
\begin{equation}
    \label{eq_prob}
    \begin{split}
     m_i &= \frac{\sum_{j=1}^{\hat{h}}{\sum_{l=1}^{\hat{w}}{\hat{M}{[i,j,l]}}}}{{\hat{h}} \cdot {\hat{w}}}, \forall i \in \{1, \dots, n\}, 
     \end{split}
\end{equation}     
\begin{equation}
    \label{eq_prob2}
    \begin{split}
     p_i &= \frac{m_i}{\sum_{j=1}^n m_j}, i \in \{1, \dots, n\}, 
     \end{split}
\end{equation}
where $m_i$ is the unnormalized gradient magnitude of the $i$-th patch, and $p_i$ is the probability of masking the $i$-th patch.

\begin{figure}[t]
\begin{center}
\centerline{\includegraphics[width=0.61\linewidth]{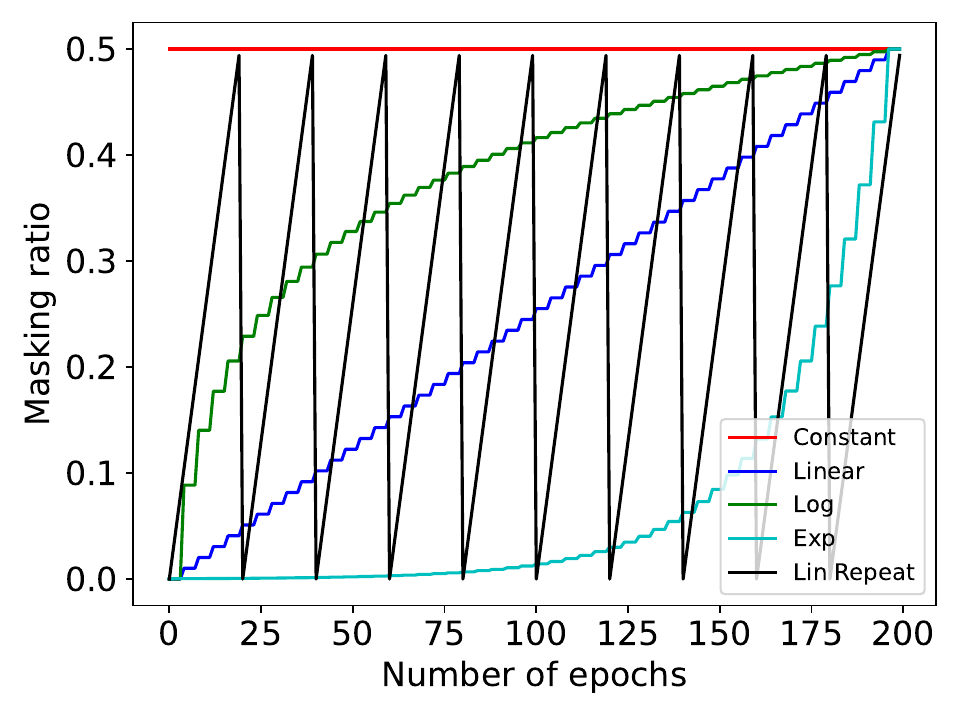}}
\caption{The proposed curriculum schedules are based on masking a certain number of image patches in each epoch. For illustration purposes, the number of epochs is $N=200$ and the maximum masking ratio is $r_N=0.5$ for all schedules. Best viewed in color.}
\label{fig:curric}
\end{center}
\end{figure}

We underline that CBM has two important hyperparameters, namely the vector $\mathbf{r}$ and the number of patches $n$. Given that the images are divided into non-overlapping patches, the number of patches $n$ is directly determined by the patch dimensions. In our experiments, we use square patches (having the same height and width). To generate $\mathbf{r}$, we consider one of the various schedules depicted in Figure \ref{fig:curric}. The \emph{constant} schedule emulates the framework proposed by He \etal~\citep{He-CVPR-2022}, which does not represent an actual curriculum, since it generates equally difficult examples during training, \ie: 
\begin{equation}
    \label{eq_const}
    r_k = r_N, \forall k \in \{1,2,...,N\},
\end{equation}
where $N$ is the number of training epochs. The \emph{constant} schedule is only added as a baseline. 

We propose three curriculum schedules that generate a \emph{logarithmic}, \emph{linear} or \emph{exponential} growth of the masking ratio. The log schedule creates a more aggressive curriculum, increasing the masking ratio much sooner:
\begin{equation}
    \label{eq_log}
    r_k = r_N \cdot \log_2 \left(1+ \frac{k}{N}\right), \forall k \in \{1,2,...,N\}.
\end{equation}
In contrast, the exp schedule is the least aggressive curriculum, introducing high masking ratios only at the very end:
\begin{equation}
    \label{eq_exp}
    r_k = r_N \cdot e^{k-N}, \forall k \in \{1,2,...,N\}.
\end{equation}
The linear schedule is between the log and exp schedule, its formula being given by:
\begin{equation}
    \label{eq_lin}
    r_k = r_N \cdot \frac{k}{N}, \forall k \in \{1,2,...,N\}.
\end{equation}


While humans learn from easy to hard, reminding easy concepts when trying to learn complex ones is generally helpful. To this end, we consider an additional curriculum learning schedule called \emph{linear repeat}, which repeats the easy-to-hard curriculum multiple times during training. This schedule is depicted in Figure \ref{fig:curric} along with the other schedules. It is repeated 10 times, at every 20 epochs, just for illustration purposes. On easy (unmasked) images, the model can learn some generic patterns from the data. As the images get harder by increasing the masking ratio, the model will try to identify new patterns that are robust to the masking procedure. In this process, the model can forget the initially learned patterns due to \emph{catastrophic forgetting}. The linear repeat scheduler helps to avoid the catastrophic forgetting of the generic patterns learned on easy images, by reintroducing easy images at various stages during the entire optimization process.

For all the aforementioned schedules, the maximum masking ratio $0 < r_N < 1$ is a hyperparameter that can be established via grid or random search on the validation set. The rate of masking can be used to control the trade-off between underfitting and overfitting. If there is no masking, the classification task becomes easy and the model can overfit the training data. In contrast, if the masking ratio is too high, the classification task can become too hard, and the model will be unable to learn from the masked data, leading to underfitting.

One of the curriculum schedules needs to be chosen before training begins. In the experiments, we present extensive results with the proposed schedules, which generate more or less aggressive curriculum learning strategies.

\section{Experiments}
\label{section_exp}

\subsection{Experimental Setup}

\noindent{\bf Data sets.} We evaluate our curriculum learning method on four object recognition data sets (CIFAR-10 \citep{Krizhevsky-TECHREP-2009}, CIFAR-100 \citep{Krizhevsky-TECHREP-2009}, ImageNet \citep{Russakovsky-IJCV-2015} and Food-101 \citep{Bossard-ECCV-2014}) and one object detection data set (PASCAL VOC \citep{Everingham-IJCV-2010}).
Each of the CIFAR-10 and CIFAR-100 data sets consists of 50,000 training images and 10,000 test images with a resolution of $32 \times 32$ pixels. CIFAR-10 contains 10 object categories, while CIFAR-100 contains 100 categories. ImageNet-1K \citep{Russakovsky-IJCV-2015} is the most popular benchmark in computer vision. We use 200 ImageNet categories in the evaluation. Food-101 \citep{Bossard-ECCV-2014} contains 75,750 training images and 25,250 test images from 101 food categories. PASCAL VOC 2007+2012 \citep{Everingham-IJCV-2010} is a well-known data set for object detection, which consists of 21,503 images. Objects from 20 categories are annotated with bounding boxes.





\noindent{\bf Architectures.} 
To evaluate the generalization capabilities of CBM across different architectures, we perform object recognition experiments with two convolutional neural networks and one transformer model. More precisely, we employ ResNet-18 \citep{He-CVPR-2016}, Wide-ResNet-50 \citep{Zagoruyko-ArXiv-2016wide} and CvT-13 \citep{Wu-ICCV-2021}. Being a transformer-based architecture, CvT benefits significantly from pre-training. Thus, we use a checkpoint pre-trained on ImageNet-21K in our experiments. As such, we apply CBM in both ``training from scratch'' and fine-tuning scenarios. Furthermore, we employ an object detection pipeline, YOLOv5 \citep{Jocher-zenodo-2022}, to evaluate the benefits of CBM in object detection. We specifically choose the YOLOv5s \citep{Jocher-zenodo-2022} model, which is pre-trained on the MS COCO data set \citep{Lin-ECCV-2014}.


\noindent{\bf Baselines.} 
We compare CBM with the conventional training regime, which uses the optimal hyperparameters (learning rate, batch size, weight decay, and so on) specific to each of the subsequent experiments. Moreover, we compare our curriculum learning approach with four competing curriculum learning methods, namely Curriculum by Smoothing (CBS) \citep{Sinha-NIPS-2020}, Label-Similarity Curriculum Learning (LSCL) \citep{Dogan-ECCV-2020}, Learning Rate Curriculum (LeRaC) \citep{Croitoru-arXiv-2022} and EfficientTrain \citep{Wang-ICCV-2023}. In the ablation study, we also compare with the framework of He \etal~\citep{He-CVPR-2022}, which can be seen as an ablated version of CBM.



\begin{table}[t]
\caption{Optimal hyperparameter configurations of the ResNet-18, Wide-ResNet-50 and CvT-13 models. For the conventional (vanilla) training regime, we report the employed optimizer, number of epochs ($N$), learning rate ($\alpha$) and mini-batch size. In addition, for the CBM regime, we report the maximum patch masking ratio ($r_N$), the number of patches per image ($n_h\times 
 n_w$), and the curriculum schedule. Note that the vector $\mathbf{r}$ is directly determined by $r_N$ and the schedule.}
\vspace{0.1cm}
\label{table:params}
\centering
\setlength\tabcolsep{2.1pt}
\begin{tabular}{|c|c|c|c|c|}
\hline
\multirow{3}{*}{\rotatebox{90}{Regime$\;\;$}} & Model & \multicolumn{2}{c|}{ResNet-18, Wide-ResNet-50} & CvT-13 \\ 
\cline{2-5}
& \multirow{3}{*}{Data set} & CIFAR-10 & \multirow{3}{*}{ImageNet} & \multirow{3}{*}{All}\\
& & CIFAR-100 & & \\
& & Food-101 & & \\
\hline
\hline
\multirow{5}{*}{\rotatebox{90}{Vanilla}} & optimizer                           & SGD     & SGD     & AdaMax  \\ 

& weight decay & $10^{-4}/5\!\cdot\!10^{-4}$ & $10^{-4}$ & -\\
& $N$                                 & 200     & 100     & 40      \\ 
& $\alpha$                            & $10^{-1}$     & $10^{-1}$     & $5\!\cdot\!10^{-4}$    \\
& mini-batch                     &   64        &   64    &  64     \\

\hline
\multirow{3}{*}{\rotatebox{90}{CBM }} & $r_N$ & $0.4/0.6$ & $0.4/0.6$ & $0.4/0.6$ \\ 
& schedule                     & Lin repeat          &  Lin repeat     &   Lin repeat    \\
& $n_h \times n_w$     & $4\times4$       & $4\times4$      & $4\times4$       \\
\hline
\end{tabular}
\end{table}




\begin{table*}[!th]
\caption{Average accuracy rates (in \%) over five runs on CIFAR-10, CIFAR-100, ImageNet and Food-101 with various neural models based on different training regimes: conventional, CBS \protect\citep{Sinha-NIPS-2020}, LSCL \protect\citep{Dogan-ECCV-2020}, LeRaC \protect\citep{Croitoru-arXiv-2022}, EfficientTrain \protect\citep{Wang-ICCV-2023}, and CBM (ours). The accuracy of the best training regime in each experiment is highlighted in bold.}\label{tab_results_recognition}
\vspace{-0.2cm}
\small{
  \begin{center}
  \begin{tabular}{|l|l|c|c|c|c|}
    \hline
    Model       & Training regime     & CIFAR-10  & CIFAR-100 & ImageNet & Food-101\\
    \hline    
    \hline
    \multirow{6}{*}{ResNet-18}        & conventional               &  $89.20 \pm 0.43$   &  $65.28 \pm 0.16$   & $57.41 \pm 0.05$ &  $68.31 \pm 0.09$ \\
              & CBS \citep{Sinha-NIPS-2020}             &  $89.53 \pm 0.22$   &  $66.41 \pm 0.21$   & $55.49 \pm 0.20$ & $65.09 \pm 0.47$\\
              & LSCL \citep{Dogan-ECCV-2020} & $ 88.28 \pm 0.14 $ & $67.59 \pm 0.25$ & $57.27 \pm 0.36$ & $69.47 \pm 0.30$\\
             & LeRaC \citep{Croitoru-arXiv-2022}    &  $89.56 \pm 0.16$   &  $66.02  \pm 0.17$   & $57.86 \pm 0.20$ & $69.57 \pm 0.07$\\
             & EfficientTrain \citep{Wang-ICCV-2023} & $89.51 \pm 0.13$ & $\mathbf{68.13} \pm 0.12$ & $57.77 \pm 0.17$ & $67.96 \pm 0,66$\\
             & CBM (ours)    &  $\mathbf{90.48} \pm 0.12$   &  $
             {67.90} \pm 0.08$   & $\mathbf{59.50} \pm 0.30$ & $\mathbf{70.77} \pm 0.23$\\
    \hline   
    \multirow{6}{*}{Wide-ResNet-50}     & conventional               &  $91.22 \pm 0.24$   &  $68.14 \pm 0.16$   & $60.97 \pm 0.30$ & $67.54 \pm 0.66$\\
         & CBS \citep{Sinha-NIPS-2020}             &  $89.05 \pm 1.00$   &  $65.73 \pm 0.36$   & $53.30 \pm 1.53$ & $58.95 \pm 1.80$ \\
         & LSCL \citep{Dogan-ECCV-2020} & $91.24 \pm 0.13$ & $68.34 \pm 0.49$ & $62.70 \pm 0.23$ & $69.20 \pm 0.48$ \\
         & LeRaC \citep{Croitoru-arXiv-2022}    &  $91.58 \pm 0.16$   &  $69.38 \pm 0.26$   & $61.98 \pm 0.42$  &  $67.96 \pm 0.35$\\
         & EfficientTrain \citep{Wang-ICCV-2023} & $91.03  \pm 0.28$ & $69.14 \pm 0.20$ & $62.44 \pm 0.11$ & $70.33 \pm 0.54$ \\
         & CBM (ours)    &  $\mathbf{92.67} \pm 0.09$   &  $\mathbf{70.90} \pm 0.31$   & $\mathbf{63.30} \pm 0.54$ & $\mathbf{72.09} \pm 0.14$\\
    \hline
    \multirow{6}{*}{CvT-13}  & conventional               &  $93.56 \pm 0.05$   &  $77.80 \pm 0.16$   & $70.71 \pm 0.35$ & $85.22 \pm 0.11$\\
      & CBS \citep{Sinha-NIPS-2020}             &  $85.85 \pm 0.15$   &  $62.35 \pm 0.48$   & $68.41 \pm 0.13$  & $81.41 \pm 0.42$\\
      & LSCL \citep{Dogan-ECCV-2020} & $93.91 \pm 0.20$ & $78.63 \pm 0.12$ & $70.96 \pm 0.30$ & $82.49 \pm 0.10$\\
     & LeRaC \citep{Croitoru-arXiv-2022}    &  $94.15 \pm 0.03$   &  $78.93 \pm 0.05$   & $71.34 \pm 0.08$ & $86.05 \pm 0.08$ \\
      & EfficientTrain \citep{Wang-ICCV-2023} & $94.50 \pm 0.17$ & $78.20 \pm 0.34$ & $71.64 \pm 0.33$ & $85.53 \pm 0.14$\\
      & CBM (ours)    &  $\mathbf{95.00} \pm 0.10$   &  $\mathbf{81.14} \pm 0.21$   & $\mathbf{71.82} \pm 0.20$ & $\mathbf{87.96} \pm 0.08$ \\
    \hline
  \end{tabular}
  \end{center}
}
\end{table*}

\noindent{\bf Hyperparameter tuning.} 
The optimal hyperparameter configurations of the ResNet-18, Wide-ResNet-50 and CvT-13 models are shown in Table \ref{table:params}. Regardless of the training regime, ResNet-18 and Wide-ResNet-50 are trained with SGD for 200 epochs on the CIFAR-10, CIFAR-100 and Food-101 data sets, and 100 epochs on ImageNet. Since CvT-13 is pre-trained on ImageNet, it only requires 40 epochs of fine-tuning to converge. The SGD optimizer is used alongside an annealing learning rate scheduler for ResNet-18 and Wide-ResNet-50, requiring a large initial learning rate of $10^{-1}$. In contrast, AdaMax does not require a scheduler, as it adjusts its learning rate based on data characteristics. We optimize all models with the cross-entropy loss. The maximum masking ratio for each model is found through grid search on the validation set, considering values between $0.1$ and $0.8$. The optimal maximum masking ratio for CIFAR-10 is $0.4$, while ImageNet seems to benefit from a higher ratio of $0.6$ when using ResNet-18 or CvT-13.
Among the proposed curriculum schedules, we select the masking ratio at each epoch based on a linear schedule, which is repeated every 5 epochs. We present results with various curriculum schedules later. The number of patches is chosen between $2\times2$ and $16\times16$. The optimal choice for all models is $4\times4$.


We perform the object detection experiments with a YOLOv5s instance that uses the CSPDarknet53 \citep{Wang-CVPRW-2020} backbone. The model is trained for $100$ epochs, using SGD with momentum. We set the learning rate to $0.01$ in all experiments and use a weight decay of $5 \cdot 10^{-4}$, with a warm-up stage of $3$ epochs, where the learning rate is increasing from $3 \cdot 10^{-6}$ to $10^{-2}$. For the LeRaC experiments, we replace the warm-up stage with the actual curriculum method. The momentum is set to $0.8$ during warm-up, and $0.94$ afterwards. The masking schedule is the same as the one employed for the recognition models, but the maximum masking ratio is $0.3$ and the number of patches is $16\times16$.

\noindent{\bf Environment.}
The experiments are conducted on a server with two Intel Xeon v4 3.0GHz CPUs, 256GB of RAM, and four Nvidia GeForce GTX 1080 GPUs, each with 11GB of VRAM.

\noindent{\bf Evaluation.}
The object recognition models are evaluated in terms of the classification accuracy. We repeat each experiment five times and report the average accuracy and the standard deviation.

For the object detection models, we compute the Average Precision (AP) on each class and report the performance over all classes, \ie~the mean Average Precision (mAP). Following standard evaluation protocols, we consider that an object is correctly detected if the intersection over union (IoU) between the predicted and ground-truth bounding boxes is at least 0.5. Hence, the reported measure is mAP@IoU=0.5.

\subsection{Results}

\noindent{\bf Object recognition.}
In Table \ref{tab_results_recognition}, we report the object recognition results with six training regimes (conventional, CBS, LSCL, LeRaC, EfficientTrain and CBM) on CIFAR-10, CIFAR-100, ImageNet and Food-101. We report the average accuracy rates and the standard deviations over five runs with each model and training regime.

On CIFAR-10, the only training regime that makes ResNet-18 surpass the $90\%$ threshold is CBM, boosting the performance by $1.28\%$ over the baseline. We observe similar gains brought by our method (CBM) to the Wide-ResNet-50 ($+1.45\%$) and CvT-13 ($+1.44\%$) architectures. In contrast, CBS is only able to bring improvements for ResNet-18, and LSCL is only able to improve CvT-13. EfficientTrain seems to be slightly better than CBS and LSCL, although it degrades the performance of Wide-ResNet-50. LeRaC is the top competitor, being consistent across all architectures, but its accuracy gains are always below $1\%$ on CIFAR-10.

On CIFAR-100, we observe that the CBS regime is not helpful for Wide-ResNet-50 and CvT-13, degrading their performance rates by significant margins. LeRaC proves to be a worthy competitor, always boosting the accuracy rate of the baseline. Its performance boosts are above $1\%$ for Wide-ResNet-50 and CvT-13. EfficientTrain is the best regime for ResNet-18, but it only ranks third and fourth for Wide-ResNet-50 and CvT-13. In contrast, CBM brings the highest accuracy gains for Wide-ResNet-50 and CvT-13, remarkably surpassing all its competitors by more than $1.52\%$ for Wide-ResNet-50 and $2.21\%$ for CvT-13, respectively.

\begin{table}[t]
\caption{Mean Average Precision (mAP) at an IoU threshold of 0.5 on PASCAL VOC 2007+2012 for the YOLOv5 \protect\citep{Jocher-zenodo-2022} model based on different training regimes: conventional, CBS \protect\citep{Sinha-NIPS-2020}, LSCL \protect\citep{Dogan-ECCV-2020}, LeRaC \protect\citep{Croitoru-arXiv-2022}, EfficientTrain \protect\citep{Wang-ICCV-2023}, and CBM (ours). The best result is highlighted in bold.}
\label{tab_obj_detection} 
\vspace{0.1cm}
\centering 
\setlength\tabcolsep{2.4pt}
\small{
\begin{tabular}{|l | c |} 
\hline
 Training regime & mAP@0.5\\
 \hline
 \hline
  conventional & $0.832 \pm 0.006$  \\
 CBS \citep{Sinha-NIPS-2020}
 & $0.829 \pm 0.003$ \\
  LSCL \citep{Dogan-ECCV-2020}
 & $0.833 \pm 0.005$ \\
 LeRac \citep{Croitoru-arXiv-2022}
 & $0.846 \pm 0.004$ \\
 EfficientTrain \citep{Wang-ICCV-2023} & $0.839 \pm 0.003$ \\
 CBM (ours) & $\mathbf{0.847} \pm 0.001$ \\
\hline
\end{tabular}
}
\end{table}

ImageNet is the most challenging benchmark included in our experiments. Still, the trends observed on CIFAR-10 and CIFAR-100 also apply to ImageNet. For example, CBS degrades the performance, just as before. In contrast, LSCL, LeRaC and EfficientTrain outperform the conventional training regime, regardless of the underlying architecture. Yet, CBM stands out as the best training regime, outperforming all competitors by more than $1.64\%$ for ResNet-18.

The results reported on Food-101 are mostly consistent with the results reported on ImageNet. CBS is often harmful, while LSCL, LeRaC and EfficientTrain are usually able to increase performance. Still, CBM exhibits the highest gains for all three architectures, showing significant differences with respect to its competitors. For instance, CBM surpasses the second-best regime for CvT-13 by $1.91\%$.


In summary, the object recognition experiments point towards some generic conclusions. The CBS \citep{Sinha-NIPS-2020} training regime is only useful in some cases. LSCL \citep{Dogan-ECCV-2020} and EfficientTrain \citep{Wang-ICCV-2023} are usually helpful, although they sometimes reduce performance. LeRaC \citep{Croitoru-arXiv-2022} brings consistent accuracy gains, but the improvements are often below $1\%$. In contrast, the proposed regime, CBM, outperforms the other training regimes on all four data sets and for all three architectures. We performed Cochran's Q statistical testing for conventional training versus CBM. The statistical tests indicate that our performance gains (on all data sets and architectures) are significant, at a p-value of $0.001$. 


\noindent{\bf Object detection.} 
We explore the applicability of the curriculum learning methods on the object detection task, reporting the mAP scores of YOLOv5s on PASCAL VOC \citep{Everingham-IJCV-2010}, while alternating between the six training regimes. The corresponding results are reported in Table~\ref{tab_obj_detection}.
The object detection results indicate that LeRaC, EfficientTrain and CBM lead to sizeable improvements, while CBS seems to degrade performance. Our method obtains an mAP of $0.847$, outperforming all other regimes. Hence, the object detection results confirm that our approach is a viable curriculum learning strategy.
 




\subsection{Ablation Results}

\begin{table}[t]
\caption{Average accuracy rates (in \%) over five runs on CIFAR-10, CIFAR-100 and ImageNet with various curriculum schedules. The accuracy of the best regime for each model and data set pair is highlighted in bold.}
\label{tab_schedules}
\centering
\setlength\tabcolsep{2.2pt}
\begin{tabular}{|c|l|c|c|c|}
\hline
Data & Curriculum & \multicolumn{3}{c|}{Model}\\
\cline{3-5}
set & schedule & ResNet-18 & WResNet-50 & CvT-13\\
 \hline
 \hline
\multirow{6}{*}{\rotatebox{90}{CIFAR-10}}     & - & 89.20$\pm$0.43 & 91.22$\pm$0.24 & 93.56$\pm$0.05 \\ 
\cline{2-5}
& Constant & 89.41$\pm$0.08 & 92.08$\pm$0.06 & 94.36$\pm$0.09 \\ 
& Linear & 90.24$\pm$0.05 & 92.39$\pm$0.11 & 94.71$\pm$0.05 \\ 
& Log & 90.31$\pm$0.09 & 92.61$\pm$0.08 & 94.49$\pm$0.07 \\ 
& Exp & 90.35$\pm$0.06 & 91.80$\pm$0.15 & 94.43$\pm$0.12 \\ 
& Lin repeat & \textbf{90.48}$\pm$0.12 & \textbf{92.67}$\pm$ 0.09 & \textbf{95.00}$\pm$0.10 \\
\cline{2-5}
\hline
\multirow{6}{*}{\rotatebox{90}{CIFAR-100}}     & - & 65.28$\pm$0.16& 68.14$\pm$0.16 & 77.80$\pm$0.16 \\ 
\cline{2-5}
& Constant & 
66.58$\pm$0.22 & 68.64$\pm$0.16 & 77.37$\pm$0.32 \\ 
& Linear & 66.46$\pm$0.28 & 70.51$\pm$0.20  & 78.94$\pm$0.25 \\ 
& Log & 67.51$\pm$0.12  & 70.75$\pm$0.30  & 79.22$\pm$0.13 \\ 
& Exp & 67.27$\pm$0.07 & 68.82$\pm$0.23 & 79.11$\pm$0.11 \\ 
& Lin repeat & \textbf{67.90}$\pm$0.08 & \textbf{70.90}$\pm$0.31 & \textbf{81.14}$\pm$0.21  \\
\cline{2-5}
\hline
\multirow{6}{*}{\rotatebox{90}{ImageNet}}     & - & 57.41$\pm$0.05 & 60.97$\pm$0.30 & 70.71$\pm$0.35 \\ 
\cline{2-5}
& Constant & 57.82$\pm$0.13 & 61.25$\pm$0.22 & 70.14$\pm$0.07 \\ 
& Linear & 58.93$\pm$0.18 & \textbf{63.61}$\pm$0.97 & 70.92$\pm$0.08 \\ 
& Log & 59.17$\pm$0.14 & 62.82$\pm$0.29 & 70.78$\pm$0.13 \\ 
& Exp & 58.02$\pm$0.23 & 62.10$\pm$0.05  & 71.35$\pm$0.14 \\ 
& Lin repeat & \textbf{59.50}$\pm$0.30 & 63.30$\pm$0.54 & \textbf{71.82}$\pm$0.20 \\
\cline{2-5}
\hline
\end{tabular}

\vspace{0.2cm}
\end{table}

\noindent{\bf Curriculum schedules.} We present additional results with various curriculum schedules in Table \ref{tab_schedules}. We include the vanilla training regime as a baseline. The constant schedule makes our framework equivalent to that of He \etal~\citep{He-CVPR-2022}. Although He \etal~\citep{He-CVPR-2022} showed that masked auto-encoders reach optimal results in self-supervised learning, we observe that their training regime leads to marginal improvements in supervised learning. However, there are multiple curriculum learning schedules that bring significant performance improvements over the vanilla training regime and the constant schedule, for each combination of model and data set. Thus, the lower accuracy rates of the constant schedule can be attributed to the lack of easy-to-hard curriculum in this type of schedule. Among the considered schedules, the \emph{linear}, \emph{log} and \emph{linear repeat} schedules exhibit the highest performance gains, but the best choice is clearly \emph{linear repeat}. This observation motivates our use of the linear repeat schedule in the experiments presented in Tables \ref{tab_results_recognition} and \ref{tab_obj_detection}.


\begin{table}[t]
\caption{Ablation study of the gradient-based masking and the curriculum schedule on CIFAR-100. We alternatively deactivate the gradient-based masking and the curriculum to determine their individual benefits.}
\label{tab_ablation} 
\centering 
\setlength\tabcolsep{3.5pt}
\small{
\begin{tabular}{|l | c | c | c |} 
\hline
 \multirow{2}{*}{Model} &
 Gradient & \multirow{2}{*}{Curriculum} & \multirow{2}{*}{Accuracy}\\
  & masking &  & \\
 \hline
\hline
 \multirow{4}{*}{ResNet-18}& \xmark & \xmark & $65.28 \pm 0.16$ \\
  & \cmark & \xmark & $66.58 \pm 0.22$\\
 & \xmark & \cmark & $67.75 \pm 0.17$\\
 & \cmark & \cmark &$\mathbf{67.90} \pm 0.08$  \\
\hline
  \multirow{4}{*}{Wide-ResNet-50} & \xmark & \xmark & $68.14 \pm 0.16$ \\
 & \cmark & \xmark & $68.64 \pm 0.16$\\
 & \xmark & \cmark & $70.45 \pm 0.11$\\
 & \cmark & \cmark & $\mathbf{70.90} \pm 0.31$\\
\hline
  \multirow{4}{*}{CvT-13} & \xmark & \xmark & $77.80 \pm 0.16$ \\
 & \cmark & \xmark & $78.37 \pm 0.32$\\
 & \xmark & \cmark & $80.02 \pm 0.16$\\
 & \cmark & \cmark & $\mathbf{81.14} \pm 0.21$\\
\hline
\end{tabular}
}  
\end{table}

\noindent{\bf Ablation of gradient masking and curriculum.} We first study the effect of our gradient-based masking procedure, replacing it with masking based on the uniform distribution of the patches. We jointly assess the influence of the curriculum schedule, as opposed to using a constant masking ratio during the whole training process. We present these experiments on CIFAR-100 in Table \ref{tab_ablation}. 
Without the curriculum, the gradient-based masking brings accuracy gains for ResNet-18 and Wide-ResNet-50, but not for CvT-13. We performed Cochran's Q statistical tests to assess the significance of gradient-based masking. For ResNet-18 and Wide-ResNet-50, the gains are significant, at p-values of $0.001$ and $0.01$, respectively. Replacing the constant schedule with the proposed curriculum schedule (linear repeat) improves the accuracy rates of all models. When the easy-to-hard curriculum is activated, the gradient-based masking increases the accuracy, but the improvements seem low. While the results indicate that both our contributions are effective, our curriculum approach has a much higher positive impact on performance. We report the highest gains when combining the gradient-based masking with our curriculum learning schedule, confirming the effectiveness of our joint design. From a practical point of view, we thus judge our contributions as relevant.

\begin{figure}[t!]
\centering
\includegraphics[width=0.7\linewidth]{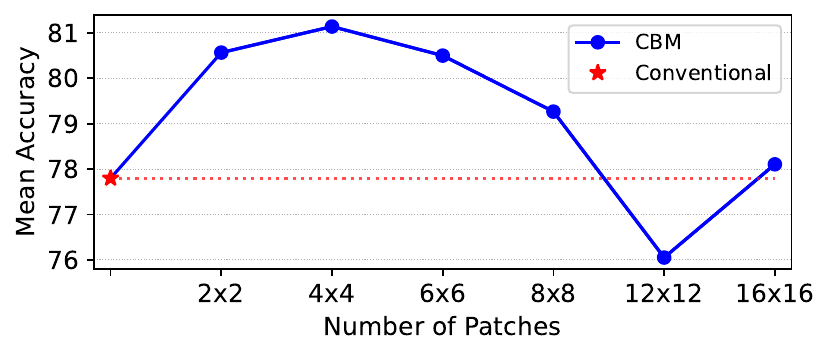}
\caption{{Varying the number of masking patches of the linear repeat schedule, for CvT-13 on CIFAR-100. There are multiple configurations that surpass the baseline.}}
\label{fig:ablation_patches}
\vspace{0.5cm}
\end{figure}

\begin{figure}[t!]
\centering
\includegraphics[width=0.7\linewidth]{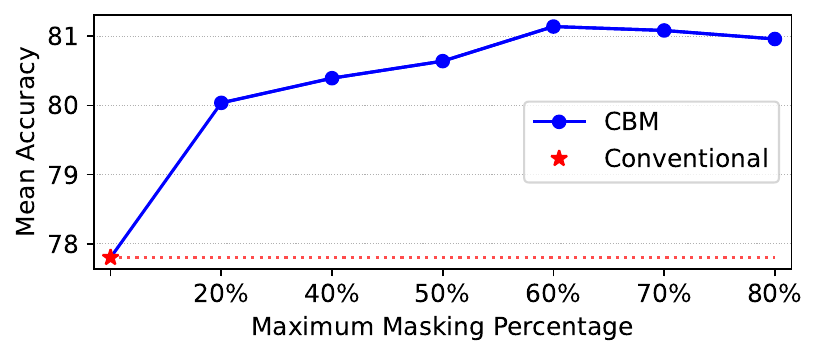}
\caption{{Varying the maximum masking ratio of the linear repeat schedule, for CvT-13 on CIFAR-100. All hyperparameter choices outperform the baseline.}}
\label{fig:ablation_mask_ratio}
\vspace{0.5cm}
\end{figure}

\begin{figure}[t!]
\centering
\includegraphics[width=0.7\linewidth]{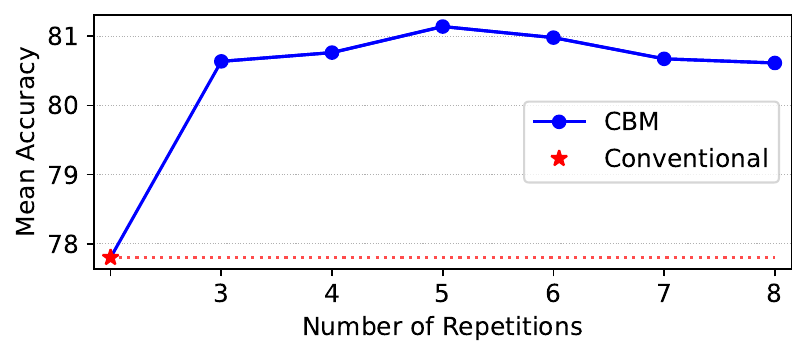}
\caption{{Varying the number of repetitions of the linear repeat schedule, for CvT-13 on CIFAR-100. All hyperparameter values lead to better results than the baseline.}}
\label{fig:ablation_repetitions}
\vspace{1.0cm}
\end{figure}

\begin{table*}[!t]
    \caption{{Results for CvT-13 on CIFAR-100 with CutMix, Masking and CutMix+Masking.}}\label{tab_results_cutmix}
    \vspace{-0.2cm}
  \begin{center}
  \setlength\tabcolsep{0.2em}
     \begin{tabular}{|l|c|c|c|c|}
    \hline
    Augmentation       & Baseline    & Constant  & Linear & Lin Repeat\\
    \hline    
    \hline
    No augmentation & $77.80 \pm 0.16$ & - & - & - \\
    CutMix         & - &  $74.94 \pm 0.52$   & $79.58 \pm 0.17$ & $80.87 \pm 0.63$ \\
    Masking        & -  &  $77.37 \pm 0.32$   & $78.94 \pm 0.25$ & $81.14 \pm 0.21$ \\
    CutMix+Masking & -  &  $79.56 \pm 0.12$   & $79.78 \pm 0.26$ & $81.34 \pm 0.25$ \\
    \hline
  \end{tabular}
  \end{center}
\end{table*}

\noindent{\bf Ablation of the linear repeat schedule.}
We conduct additional ablation studies for our best schedule, namely linear repeat. In Figure \ref{fig:ablation_patches}, we present results with CvT-13 \cite{Wu-ICCV-2021} on CIFAR-100 \cite{Krizhevsky-TECHREP-2009} while varying the number of patches for the linear repeat schedule. We observe that any number of patches between $2 \times 2$ and $8\times 8$ leads to better results than the baseline. This confirms that our approach is fairly robust to suboptimal configurations of the number of patches. 
In Figure \ref{fig:ablation_mask_ratio}, we show results with CvT-13 \cite{Wu-ICCV-2021} on CIFAR-100 \cite{Krizhevsky-TECHREP-2009} while varying the maximum masking ratio for the linear repeat schedule. Remarkably, all configurations produce better results than the baseline. However, we generally observe larger improvements for higher maximum masking ratios.
In Figure \ref{fig:ablation_repetitions}, we present additional results on CIFAR-100 \cite{Krizhevsky-TECHREP-2009} with linear repeat, using different repeat intervals. Once again, we notice that all hyperparameter choices outperform the baseline.
In summary, the ablation results show that CBM based on linear repeat is very robust to suboptimal hyperparameter choices.

\subsection{Additional Results}

\noindent{\bf Augmentation versus curriculum.}
We underline that the constant schedule is equivalent to simple data augmentation via masking patches. As shown in Table 4 from the main article, the differences between the baseline and the constant schedule (which correspond to data augmentation) and the differences between the constant and linear repeat schedules show that curriculum learning brings higher improvements on CIFAR-100 and ImageNet. Moreover, simple data augmentation degrades performance for CvT-13 on CIFAR-100 and ImageNet. This confirms that the reported improvements are mostly due to curriculum learning.

To further show that our curriculum strategy can be applied to other data augmentations, we conduct experiments with CutMix \cite{Yun-ICCV-2019} and curriculum learning for CvT-13 on CIFAR-100. As shown in Table \ref{tab_results_cutmix}, the CutMix augmentation does not work at all without the curriculum learning strategies, namely linear and linear repeat. Notably, combining CutMix and masking with the linear repeat curriculum leads to additional gains. These results support our previous observation, namely that the performance gains are rather the effect of employing curriculum learning.

\begin{figure}[!t]
\begin{center}
\centerline{\includegraphics[width=0.47\linewidth]{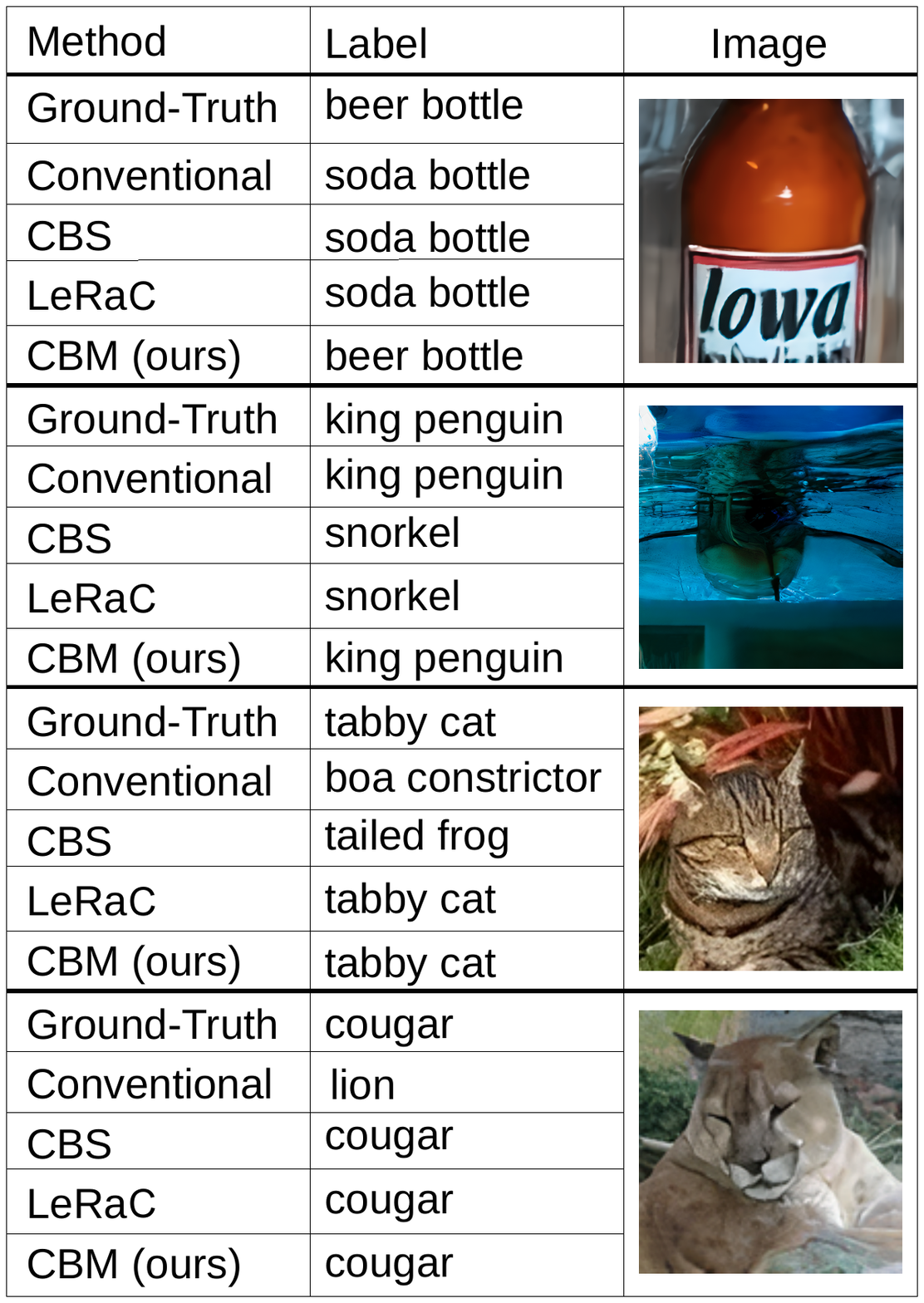}}
 \vspace{0.06cm}
 \caption{Qualitative results for the image recognition task, comparing the vanilla regime with CBS \protect\cite{Sinha-NIPS-2020}, LeRaC \protect\cite{Croitoru-arXiv-2022}, and CBM (ours). The samples are taken from ImageNet \protect\cite{Russakovsky-IJCV-2015}. In the illustrated examples, we can observe that our curriculum learning method helps the model to better distinguish between similar classes. Best viewed in color.}
\label{fig_qualitative_rec}
\end{center}
\end{figure}


\begin{table}[!t]
    \caption{Accuracy rates (in \%) on CIFAR-10 with the ViT model based on two training regimes: conventional and CBM.}\label{tab_vit_results}
      \vspace{-0.2cm}
  \begin{center}
     \begin{tabular}{|l|c|c|c|c|}
    \hline
    Data set       & ViT (conventional)   & ViT (CBM) \\
    \hline    
    CIFAR-10    &  $94.55 \pm 0.16$ & $95.32 \pm 0.21 $ \\
    \hline
  \end{tabular}
  \end{center}
\end{table}

\noindent{\bf Compatibility with vision transformers.}
In Table \ref{tab_results_recognition}, we report results with CvT for a direct comparison with LeRaC \citep{Croitoru-arXiv-2022}, one of our main competitors, which also uses CvT. However, our method is not tied to a certain architecture. To demonstrate this, we conduct additional experiments with ViT \citep{Dosovitskiy-ICLR-2020} on CIFAR-10. The results shown in Table \ref{tab_vit_results} confirm that CBM is also useful for ViT.

\noindent{\bf Qualitative results.}
In Figure~\ref{fig_qualitative_rec}, we include a few examples to illustrate the benefits brought by our curriculum learning method on the classification task. We can see that the model trained with our strategy is able to better discriminate between similar classes. Indeed, the first three image samples show that our method outperforms the other ones in the shown cases. As for the last example in Figure~\ref{fig_qualitative_rec}, CBM aligns with the other curriculum learning strategies and predicts the correct class for the corresponding image.

\section{Conclusion}

In this paper, we proposed a new curriculum learning method based on masking input patches. Our method uses a gradient-based masking procedure in conjunction with a gradually increasing masking ratio to create an easy-to-hard curriculum without having to estimate the difficulty of training samples. To demonstrate the effectiveness of CBM, we carried out a comprehensive set of experiments, considering multiple neural architectures and data sets. Our empirical results show that CBM surpasses the vanilla training regime, as well as four state-of-the-art curriculum learning strategies \citep{Croitoru-arXiv-2022,Dogan-ECCV-2020,Sinha-NIPS-2020,Wang-ICCV-2023}. Moreover, we performed ablation experiments to justify our design integrating the gradient-based masking and the easy-to-hard curriculum.






\bibliography{bibliography}

\section{Supplementary}

In the supplementary, we present more qualitative results to further assess the performance of the proposed curriculum learning strategy in object detection. Additionally, we provide an empirical comparison with CL-MAE.

\begin{figure*}[!tbh]
\begin{center}
\centerline{\includegraphics[width=1.0\linewidth]{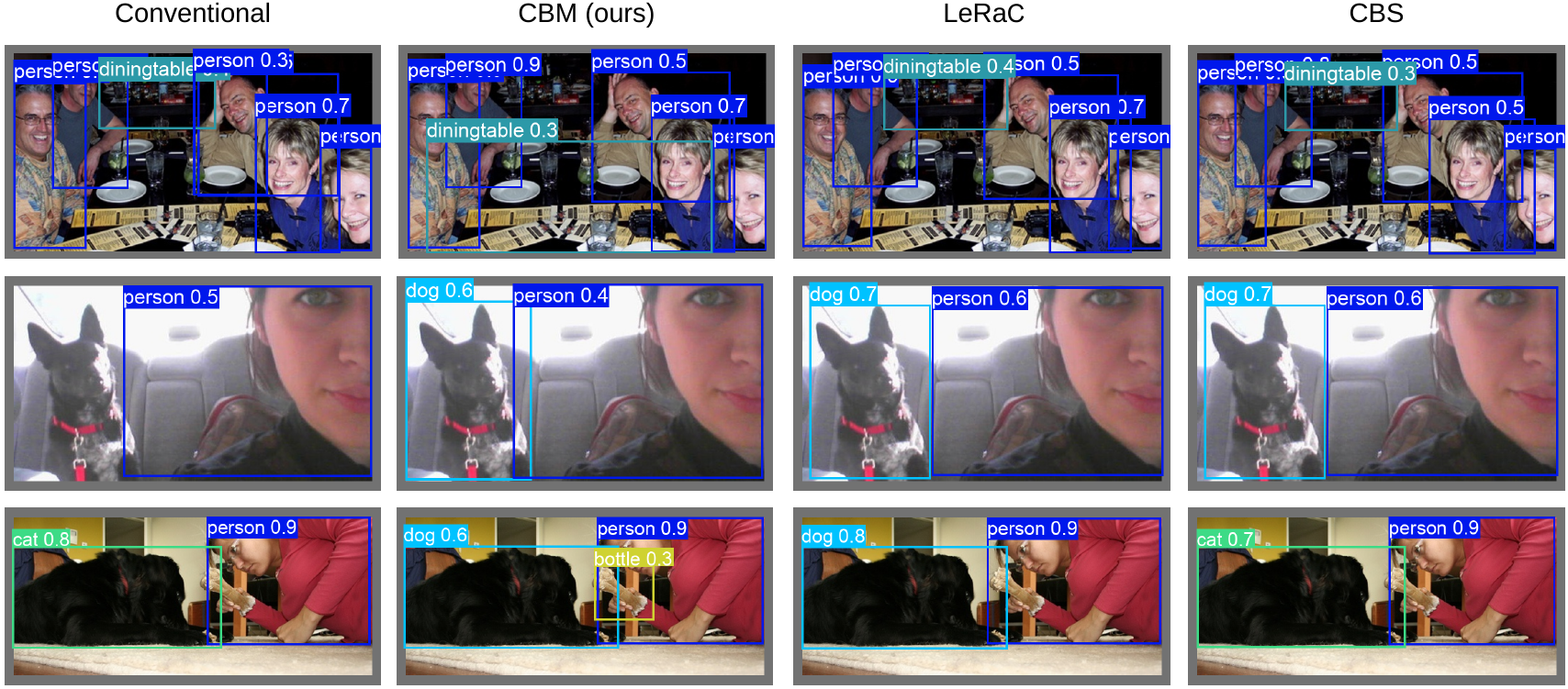}}
 \vspace{0.1cm}
 \caption{Qualitative results for object detection on PASCAL VOC 2012 \protect\cite{Everingham-IJCV-2010}, comparing the vanilla regime with CBS \protect\cite{Sinha-NIPS-2020}, LeRaC \protect\cite{Croitoru-arXiv-2022}, and CBM (ours). In the first image, CBM is the only method that finds the dining table correctly. In the second image, CBM aligns with the other curriculum learning methods and identifies the dog. In the third example, CBM classifies correctly the dog, but also mistakes a bone with a bottle. Best viewed in color.}
\label{fig_qualitative_det}
\end{center}
\end{figure*}

\subsection{Additional Qualitative Results}

Figure~\ref{fig_qualitative_det} includes qualitative results for the object detection task. We observe that CBM performs better than the baseline in all three examples. Moreover, in the first example, CBM is the only method that correctly identifies the dining table.


\begin{table*}[t]
\caption{Accuracy rates for the top-1 and top-5 results on the Architectural Heritage Elements \protect\citep{AHE-2021}, Sea Animals \protect\citep{SeaAnimals-2024} and Sport Balls \protect\citep{SportsBalls-2022} data sets for CL-MAE \protect\citep{Madan-WACV-2024} versus CBM (ours). The best results are highlighted in bold.}
\label{tab_vs_clmae} 
\vspace{0.1cm}
\centering 
\setlength\tabcolsep{2.4pt}
\small{
\begin{tabular}{|l | c | c | c | c | c | c |} 
\hline
\multirow{2}{*}{Training regime}    & \multicolumn{2}{c|}{Architectural Heritage Elements} & \multicolumn{2}{c|}{Sea Animals}  & \multicolumn{2}{c|}{Sport Balls} \\
\cline{2-7}
 & Accuracy@1 & Accuracy@5 & Accuracy@1 & Accuracy@5 & Accuracy@1 & Accuracy@5 \\
 \hline
 \hline
  CL-MAE \cite{Madan-WACV-2024} & $86.8$ & $99.9$ & $67.9$ & $92.7$  & $65.8$ & $90.7$ \\
 CBM (ours) & $\mathbf{97.15} \pm 0.11$ & $\mathbf{100.00} \pm 0.00$ & $\mathbf{89.91} \pm 0.53$ & $\mathbf{98.00} \pm 0.08$ & $\mathbf{95.44} \pm 0.06$ & $\mathbf{99.31} \pm 0.07$ \\
\hline
\end{tabular}
}
\end{table*}

\subsection{Comparison with CL-MAE}

CL-MAE \cite{Madan-WACV-2024} employs curriculum learning during the pre-training stage of a masked auto-encoder, whereas CBM is used either for fine-tuning or for training models from scratch. Additionally, CL-MAE relies on a learnable module to determine which patches to mask, while keeping the masking ratio constant during training. In contrast, our method utilizes image gradients to mask the salient regions and leverages progressively higher masking ratios to create the curriculum. Moreover, CBM is independent of the underlying network architecture (shown through experiments with various architectures, \eg~ResNet and CvT), while CL-MAE is specifically designed for masked auto-encoders.

To provide a quantitative comparison between our method and CL-MAE \cite{Madan-WACV-2024}, we conduct additional experiments, applying CBM to the CvT architecture on the Architectural Heritage Elements \cite{AHE-2021}, Sea Animals \cite{SeaAnimals-2024} and Sports Ball \cite{SportsBalls-2022} datasets, which are used in the CL-MAE paper \cite{Madan-WACV-2024}. The results of these experiments are shown in Table \ref{tab_vs_clmae}. While CBM produces significantly better results, we acknowledge that the training regimes of CL-MAE (pre-training and linear probing) and CBM (fine-tuning) are different.

\end{document}